# REET: Robustness Evaluation and Enhancement Toolbox for Computational Pathology

Alex Foote[$], Amina Asif*,[$], Nasir Rajpoot and Fayyaz Minhas*

Tissue Image Analytics Centre, Department of Computer Science, University of Warwick, UK.

*To whom correspondence should be addressed.

[$]Joint first authorship

**Abstract**

**Motivation:** Digitization of pathology laboratories through digital slide scanners and advances in deep learning approaches for objective histological assessment have resulted in rapid progress in the field of computational pathology (CPath) with wide-ranging applications in medical and pharmaceutical research as well as clinical workflows. However, the estimation of robustness of CPath models to variations in input images is an open problem with a significant impact on the downstream practical applicability, deployment and acceptability of these approaches. Furthermore, development of domain-specific strategies for enhancement of robustness of such models is of prime importance as well.

**Implementation and Availability:** In this work, we propose the first domain-specific Robustness Evaluation and Enhancement Toolbox (REET) for computational pathology applications. It provides a suite of algorithmic strategies for enabling robustness assessment of predictive models with respect to specialized image transformations such as staining, compression, focusing, blurring, changes in spatial resolution, brightness variations, geometric changes as well as pixel-level adversarial perturbations. Furthermore, REET also enables efficient and robust training of deep learning pipelines in computational pathology. REET is implemented in Python and is available at the following URL: https://github.com/alexjfoote/reetoolbox.

**Contact:** Fayyaz.minhas@warwick.ac.uk

## 1 Introduction

Computational Pathology (CPath) deals with modelling and analysis of digitized images of tissue slides. The area has seen a major surge in application of deep learning (DL) for modeling prediction tasks like tissue and cell classification, transcriptomic expression and mutation prediction, risk stratification and survival analysis (Pronier *et al.*, 2020; Lu *et al.*, 2021; Skrede *et al.*, 2020; Kather *et al.*, 2020; Fu *et al.*, 2020; Bilal *et al.*, 2021). Hundreds of recent studies using different flavors of DL have been published reporting high accuracies over different tasks using cross-validation and independent testing sets (Rakha *et al.*, 2021; Srinidhi *et al.*, 2021; Acs *et al.*, 2020; van der Laak *et al.*, 2021; Sultan *et al.*, 2020; Steiner *et al.*, 2021). However, recent work by our group as well as other researchers has shown that even highly accurate CPath models can be very fragile to small and visually imperceptible perturbations in input images (Foote *et al.*, 2021; Schömig-Markiefka *et al.*, 2021). Lack of robustness to small image perturbations – which can arise from differences in slide preparation and staining, scanning and compression pipelines – poses a major challenge to the deployment and practical use of CPath models as well as their wider acceptability. Accuracy based assessment of CPath predictors, though necessary, is not sufficient for capturing their true generalization performance and systematic evaluation of robustness to input image perturbations is needed for evaluating DL models in CPath. In this paper, we present the first Robustness Evaluation and Enhancement Toolbox (REET) for CPath models with twofold objectives: first, it enables large-scale robustness analysis; second, it provides efficient strategies for robust training (or *robustification*) of CPath pipelines.

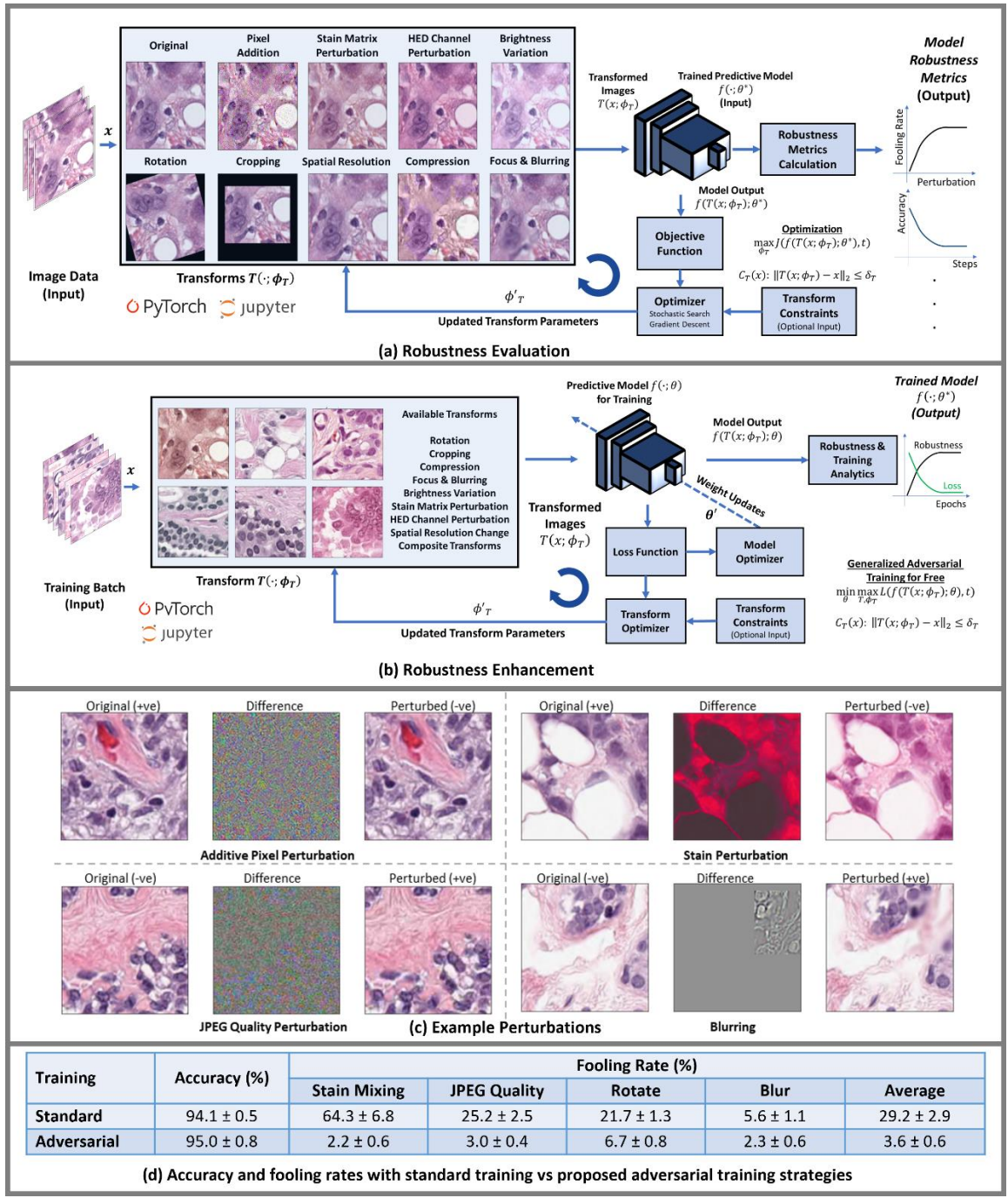

Figure 1 (a) Overall flow of robustness evaluation component. The module accepts a dataset of tissue images and a trained model to evaluate the impact of different CPath-specific perturbations on the model's accuracy and robustness. (b) Robustification component of REET. This module accepts an untrained model and a training set and performs adversarial training for free (ATF) using different transforms. (c) Example positive and negative images before and after perturbations generated by the evaluation module of REET. The predicted labels are flipped after perturbation demonstrating fragility of the model to these variations. (d) Experiment results: The first column presents accuracy of a CPath tumor detection model using the standard training and ATF implementation in REET. The rest of the columns show fooling rates to different perturbations implemented in REET. A significant improvement can be seen in terms of fooling rates after using our adversarial training.

## 2  Design and Features

REET is a Python library for robustness quantification of CPath models and robustification against several CPath-specific image variations. **Figure 1a and 1b** show the overall workflow and internal design of the robustness evaluation and enhancement components of REET. Further details including installation guidelines, example notebooks as well as details on the internal architecture and experimental validation of the proposed toolbox are available in the online repository and supplementary material.

## 2.1 Robustness Evaluation

The most common sources of variation in digitized whole-slide images (WSIs) of tissue slides are: different dye compositions in staining, slight differences in protocols for slide preparation across different pathology laboratories or in their following, variations in spatial resolution, brightness, contrast and compression due to different scanning technologies as well as imaging artefacts such as pixel noise, blurring and focusing issues. The evaluation module of REET allows robustness assessment of a predictive model by essentially simulating different types of such perturbations to input images and analyzing their effects on model output. REET takes a trained DL model and an associated test dataset comprising tissue images as input and allows the user to quantitatively assess the fragility of the given model to a series of domain-specific transformations over images in the test set in the form of metrics such as change in accuracy, *fooling rate* and visual results (see **Figure 1a**). REET also allows the user to implement their own robustness evaluation metrics as well as custom transformations and pipelines.

REET computes the sensitivity of model output with respect to changes in transform parameters that control the impact of different transforms on input images. To generate transformed images, REET uses adversarial constrained optimization of transform parameters resulting in automated identification of failure modes of the given model. **Figures 1c and 1d** show the results of robustness evaluation analysis for a tumor cell prediction model trained to predict if a given tissue image contains tumor cells (+ve label) or not (–ve label). As can be seen in the figure, different perturbations in the input image can change the prediction label generated by the model even though the resulting change between images is quite small. Fooling rate calculations for the model to different types of perturbations, as in the top row of **Figure 1d**, show that the model is very sensitive to stain variations.

REET differs from conventional adversarial attack toolboxes for natural images in that it supports domain-specific transformations such as stain changes, compression, focusing/blurring and spatial resolution variations in addition to conventional additive pixel perturbations. REET provides both projected gradient descent and stochastic search for optimization of transform parameters, useful for robustness assessment of models without requiring access to the internal architecture or weights.

## 2.2 Robustness Enhancement

The overall flow of robustness enhancement using REET is shown in **Figure 1b**. REET works by using a transform optimizer to identify transform parameters that can cause significant changes in the output of the model while minimizing the model loss function through a model optimizer. In addition to robustness evaluation, REET also supports robust training of CPath models. For this purpose, the toolbox provides a novel generalized implementation of *adversarial training for free (ATF)* (Shafahi *et al.*, 2019). Conventional adversarial training strategies have been shown to improve robustness in medical imaging models (Apostolidis and Papakostas, 2021), however, they are computationally expensive. ATF has been shown to achieve similar robustness as the conventional techniques without significantly increasing the computational cost. The original implementation of ATF allows for training using additive pixel perturbations only. The ATF implementation in REET is generalized in the sense that it can be used to train a given CPath model with an extensive array of domain-specific transformations. For this purpose, REET accepts an untrained deep learning model and a training dataset and implements an efficient ATF loop which results in improved robustness of the predictive model. The robustness of the trained model can then be evaluated over a test or validation set. **Figure 1d** shows improvement in fooling rates of a tumor cell detection model through the use of the robustness enhancement component of REET. In addition to a minor improvement in terms of accuracy in the enhanced model, we observed a significant drop in fooling rates to different domain-specific perturbations, demonstrating the effectiveness of the toolbox in robustification of models.

## 3 Conclusions

The proposed REET toolbox fills an important gap in the rapidly emerging field of computational pathology by providing a modular framework for fragility evaluation and robustification of deep learning pipelines in this domain.

## Funding

AA is funded by NIHR HTA grant (award DS406118). FM and NR are partly funded by PathLAKE consortium (https://www.pathlake.org/).

*Conflict of Interest:* NR and FM are in receipt of research funding from GSK. NR is co-founder and CSO of Histofy Ltd.